\documentclass{article}
\usepackage{spconf,amsmath,graphicx, bm, amssymb, hhline}
\usepackage{array,multirow}
\usepackage[dvipsnames]{xcolor}
\usepackage{soul}

\definecolor{green}{rgb}{0.0, 0.5, 0.0}

\newcommand\blfootnote[1]{
    \begingroup
    \renewcommand\thefootnote{}\footnote{#1}
    \addtocounter{footnote}{-1}
    \endgroup
}

\title{Augmenting Conformers with structured state-space sequence models for online speech recognition}
\name{\begin{tabular}{c}Haozhe Shan\textsuperscript{1\dag} \quad  Albert Gu\textsuperscript{2\dag}\quad  Zhong Meng\textsuperscript{3}\quad  Weiran Wang\textsuperscript{3} \quad  \\
			Krzysztof Choromanski\textsuperscript{3}\quad Tara Sainath\textsuperscript{3}\end{tabular}}

\address{\textsuperscript{*} Harvard University \\
\texttt{shanhaozhe@gmail.com} \\
\textsuperscript{2} Carnegie Mellon University \\
\texttt{agu@cs.cmu.edu} \\
\textsuperscript{3} 
Google LLC.\\
\texttt{\{zhongmeng,weiranwang,kchoro,tsainath\}@google.com}}
\begin{document}
\ninept
\maketitle
\blfootnote{* Work done while at Google.}
\begin{abstract}
Online speech recognition, where the model only accesses context to the left, is an important and challenging use case for ASR systems. In this work, we investigate augmenting neural encoders for online ASR by incorporating structured state-space sequence models (S4), a family of models that provide a parameter-efficient way of accessing arbitrarily long left context. We performed systematic ablation studies to compare variants of S4 models and propose two novel approaches that combine them with convolutions. We found that the most effective design is to stack a small S4 using real-valued recurrent weights with a local convolution, allowing them to work complementarily. Our best model achieves WERs of 4.01\%/8.53\% on test sets from Librispeech, outperforming Conformers with extensively tuned convolution.

\end{abstract}

\maketitle

\section{Introduction}
\label{sec:intro}
\vspace*{-1ex}

The search of neural architectures for encoding audio features in automatic speech recognition (ASR) has been an important and ongoing research direction. The commonly used building blocks are deep neural networks (DNN)~\cite{hinton2012deep}, recurrent neural networks (RNN) such as LSTMs~\cite{lstm1997}, convolutional neural networks (CNN) ~\cite{lecun1998}, and multi-head self attention (MHSA)  ~\cite{attention2017}. These components have different inductive biases for learning and characteristics for optimization. The recent trend is to leverage capabilities of different components and carefully combine them. For example, \cite{gulati2020conformer} and \cite{peng2022branchformer} proposed combining the global context as modeled by MHSA, and local context as modeled by depthwise convolution (simply ``convolution" afterwards), to build state-of-the-art encoders for ASR, i.e., Conformers.

Recently, structured state-space sequence models (S4) \cite{gu2021efficiently,gupta2022diagonal,gu2022parameterization,ma2023mega,smith2023simplified} emerged as a promising approach to sequence modeling. These models can be variably interpreted as  RNNs, CNNs, or classical state-space models (SSM)~\cite{gu2021combining}. As sequence models, S4s have been shown to have advantages such as being able to capture long-term dependencies and enjoy time complexity sub-quadratic in sequence length. Naturally, this raises the question of whether they can be used to augment ASR encoders. There have been several studies in this direction. \cite{fathullah2023multihead} proposed an attention-free multi-head state-space (MH-SSM) architecture that achieves competitive performance on offline ASR without MHSA. Motivated by the CNN view of S4s, \cite{ibmssm2023} introduced the DSSformer for offline ASR, where convolution components in the Conformer encoder are replaced with S4s. Besides these encoder-focused studies, \cite{ssmdecoder2023} used S4s in the attention-based \textit{decoder}, where the per-layer MHSA is replaced with S4s while the cross-attention remains unchanged. All such work focused on offline ASR where encoders have full context.

\noindent\textbf{Our contributions} In this paper, we investigate augmenting ASR encoders with S4s in both the offline setting, similar to \cite{fathullah2023multihead} and \cite{ibmssm2023}, and the online setting, which was not covered by previous studies. We systematically compared the approach from \cite{fathullah2023multihead}, where the S4 is used as a drop-in replacement (DIR) for convolution in Conformers, with two novel approaches: one where the S4 is stacked with a local convolution component (COM); another where the S4 is used to reparameterized the finite-size kernel within the convolution component (REP). In all three approaches, we present careful ablation studies on various S4 design choices. We found that in online ASR, the DIR approach produces word-error rate (WER) on par with but no better that that from conformers. On the other hand, both novel approaches that we propose produce small but consistent WER reduction. In both online and offline settings, we found that COM is the most effective approach. In addition, success of the REP approach suggests that S4s can be effective even when they are forced to model local context, pointing to the need of further theoretical understanding of their capabilities.


\begin{figure*}[t]
    \centering
    \includegraphics[width=1.0\textwidth]{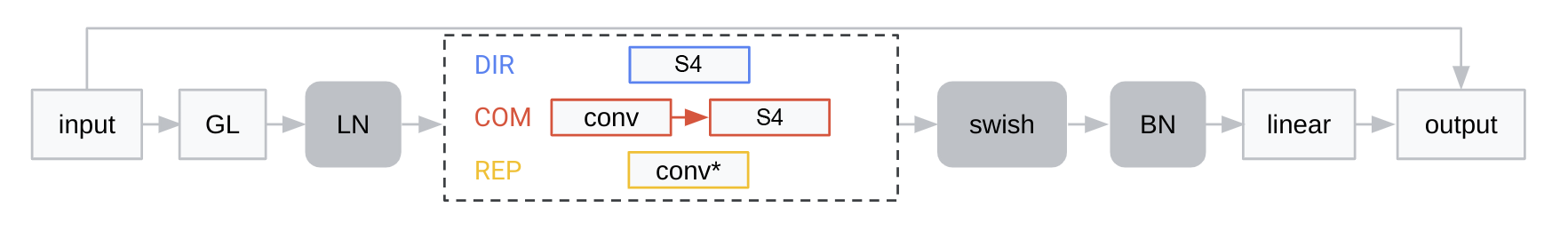}
    \caption{Architecture of the augmented convolution module. GL=gated linear ; LN=layer normalization; conv=convolution; BN=batch normalization. Inside the dashed box are three approaches of using S4s that we studied. DIR=drop-in replacement, COM=combination, REP=reparameterization. Following conventions from \cite{gulati2020conformer}, the entire block shown here is referred to as the convolution \textit{module}, while we study ways to modify the convolution \textit{component} within it.}
    \vspace*{-1ex}
    \label{fig:schematics}
\end{figure*}

\section{Structured state space sequence models}
\label{sec:format}
\vspace*{-1ex}

In this work, we exclusively considered the S4D variants of S4 models \cite{gu2022parameterization}, which have been shown to be simpler to implement and cheaper to compute while being as effective as older variants. We refer the readers to \cite{gu2022parameterization} for further details and summarize the structure here. Each model has a dimension $N$, which can be viewed as the hidden-state size. It contains 4 sets of trainable parameters ($H$ denotes the input dimension): $N\times N$ weights $\bm{A}$, $N\times H$ weights $\bm{B}$, $H \times N$ weights $\bm{C}$, and $H\times H$ residual weights $\bm{D}$. Following \cite{gu2022parameterization}, $\bm{B}$ are assumed to be uniformly $1$ and untrainable. For mapping a discrete-time input $\{\bm{u}_k \in \mathbb{R}^H\}_{k=1,...,T}$ to a discrete-time output $\{\bm{y}_k \in \mathbb{R}^H\}_{k=1,...,T}$ with a timestep $\Delta$, this model is discretized via the zero-order hold method \cite{gupta2022diagonal} as
\begin{align}
    \bm{x}_k &= \bar{\bm{A}} \bm{x}_{k-1} + \bar{\bm{B}} \bm{u}_k \quad \quad \bm{y}_k = \bm{C} \bm{x}_k + \bm{D} \bm{u}_k \nonumber \\
    & \bar{\bm{A}} = e^{\bm{A} \Delta}\quad \quad \bar{\bm{B}} = (\bar{\bm{A}}-\mathbb{I}) \bm{A}^{-1} \bm{B}.
    \label{eq:simple S4}
\end{align}
These expressions reveal that computationally, S4s can be seen as specially parameterized linear RNNs with recurrent weights $\bar{\bm{A}} \in \mathbb{C}^{N \times N}$, input weights $\bar{\bm{B}} \in \mathbb{C}^{N \times H}$, readout weights $\bar{\bm{C}} \in \mathbb{C}^{H \times N}$, and residual weights $\bm{D}\in \mathbb{R}^{H\times H}$. This allows their efficient computation via scan algorithms \cite{martin2017parallelizing}. Via the well-known duality between linear RNNs and time convolution, their output $\{ \bm{y}_k \}$ (before adding the residual part) is equivalently the result of causally convolving the input sequence with a size $T\times H$ kernel, given by
\begin{equation}
    \bar{\bm{K}} = [\bm{C} \bar{\bm{B}}, \bm{C} \bar{\bm{A}} \bar{\bm{B}}, ..., \bm{C} \bar{\bm{A}}^{T-1} \bar{\bm{B}}].
\end{equation}
Note that $\bar{\bm{K}}$ can be computed for input sequences of arbitrary length using a fixed number of parameters. 

Recent interest in applying S4s to sequence modeling arises from a series of work that proposed effective ways to parameterize and initialize such models \cite{gu2021efficiently, gupta2022diagonal, gu2022parameterization}. In particular, \cite{gu2022parameterization} suggested restricting the $\bm{A}$ to the space of \textit{diagonal} matrices for computational efficiency. In addition, to ensure that powers of $\bar{\bm{A}}$ does not diverge and lead to exploding gradients, it is sufficient to ensure that $\bm{A}$ has eigenvalues with negative real parts. We considered two parameterization-initialization schemes from \cite{gu2022parameterization}. In the \textbf{S4D-Lin} scheme, $\bm{A}$ is a complex diagonal matrix of size $N \times N$ (it thus contains $2N$ trainable parameters, composed of real and imaginary components of the diagonal entries). The $n$-th diagonal entry is initialized at $A_n^{lin} = -\frac{1}{2} + i \pi n \quad (n=0,1,...,N-1)$.
Both real and imaginary parts are trainable in this scheme. Real parts are ensured negative by reparameterizing them with a non-positive function (e.g., $Re(A_n)=-\exp(x_n)$, where $x_n$ is trained). In the \textbf{S4D-Real} scheme, $\bm{A}$ is a real diagonal matrix (with $N$ trainable parameters) with the $n$-th diagonal entry initialized at $A_n^{real} = -n-1$
and again ensured negative via reparameterization.

We utilize S4s as parts in a sequence-to-sequence encoder. During training and inference, an input of size $B \times T \times H$ is passed to the S4, where $B$ is the batch size. 
We split the input along the feature dimension into as $H$ one-dimensional time series,  and each passes through an S4. Each of the $H$ S4s 
is parameterized by the 3-tuple $(\text{diag}(\bm{A}^{(h)}),\bm{B}^{(h)},\bm{C}^{(h)})_{h=1,...,H}$, and we assume $\bm{A}^{(h)}$ to be tied across $h=1,...,H$ (while $B^{(h)}$, $C^{(h)}$ are not). As a concrete example of model size, for encoder dimension $H=512$ and $N=4$, these S4s collectively have around 4000 parameters.

\section{S4-augmented ASR encoders}
\vspace*{-1ex}

The Conformer \cite{gulati2020conformer} is the current state-of-the-art encoder architecture for ASR. The baseline Conformer encoder that we studied uses hyperparameter settings of the Conformer (L) encoder from \cite{gulati2020conformer}. Each Conformer layer consists of a MHSA module that is efficient at capturing long-term dependency and a convolution module that is good at capturing local dependency. Previous research on augmenting offline ASR encoders with S4s \cite{fathullah2023multihead} indicates that MHSA modules are indispensable for state-of-the-art performance. Therefore, we did not modify the MHSA module and restrict our analysis to augmenting the convolution module with S4s. Concretely, we considered three approaches (Fig. 1)

\begin{itemize}
    \item \textit{Drop-in Replacement} (DIR). Here, the convolution component inside the module is replaced by an S4. As the S4 has unlimited left context, this endows the modified module with unlimited left context. We note that this is the approach taken in \cite{gupta2022diagonal} for offline ASR, albeit with older S4 variants.

    \item \textit{Combination} (COM). Recent work utilizing S4s for sequence modeling has suggested a potential advantage of combining the S4 with a local (small kernel size) convolution \cite{poli2023hyena, dao2022hungry}. We therefore propose stacking them inside the module. This approach similarly endows the module with unlimited left context.

    \item \textit{Reparameterization} (REP). As discussed above, the S4 can be viewed as a way to parameterize a convolution kernel, $\tilde{\bm{K}}$, that can be match the length of arbitrary inputs. This perspective suggests that, similar to how Conformers perform better when their convolution kernel size is tuned to some finite value, performance of the S4 can be similarly improved by truncating $\bm{\tilde{K}}$ along the time dimension. This essentially means that we are using Conformers but the (finite size) convolution kernel is reparameterized by the S4. Unlike the previous two approaches, this does not increase the left context size of the module compared to Conformers.
\end{itemize}
All approaches only modify the convolution component inside the convolution module. All other settings are the same as those used in the baseline Conformer. The same modification is applied to all layers in the network. The resultant encoder models are collectively referred to as S4formers. 

\section{Experiments}
\vspace*{-1ex}

To compare the various S4formers as well as the baseline Conformer, we performed both offline and online ASR on the Librispeech dataset \cite{panayotov2015librispeech}.  For the decoder, we used the RNN-Transducer model \cite{graves2012sequence} equipped with a single-layer LSTM decoder for all experiments; label-sync and frame-sync beam search with beam size $8$ was used for decoding in the offline and online settings respectively. We computed 80-channel filterbank features from input audio and apply two layers of 2D-convolution subsampling~\cite{baevski2020wav2vec}. The resulted audio features, at a frame rate of 25Hz, are fed to the encoder. Models were trained on all 970 hours of labeled training set. SpecAugment~\cite{park2019specaugment} and variational noise~\cite{graves2011practical} were used to control overfitting.
For the baseline Conformer, we used the recommended configuration with around 119M parameters from 
\cite{gulati2020conformer}, with 17 layers, 8 attention heads, convolution kernel size 32, encoder dimension 512,  and relative positional embedding \cite{dai2019transformer}. In the offline scenario, the attention module has unlimited left and right context. In the online scenario, the attention module only accesses unlimited left context; right context was also removed from the wav2vec convolution and all convolution modules (so the effective filter size becomes 16). Hyperparameter tuning and early stopping were done based on the average WER on dev-clean and dev-other. 

\subsection{Offline ASR with S4formers}
\vspace*{-1ex}

\begin{table}[t]
    \centering
        \caption{WER (\%) of offline ASR using encoders augmented with DIR/COM approaches. All S4formers used S4s with the S4D-Real scheme.}
    \label{table: offline}
    \begin{tabular}{|c|c|c|c|c|c||c|}
    \hline
         \multirow{2}{*}{Approach} & \multirow{2}{*}{$N$} & dev & dev & test & test & test  \\
          & & clean & other & clean & other & avg.\\
         \hline 
         Conformer & NA & 1.84 & 4.49 & 2.12 &4.63 & 3.38 \\
         \hline
         \multirow{2}{*}{DIR} & 2 & 1.87 & 4.43 & 2.08 &4.61 & 3.35\\

           & 4 & 1.89 & 4.54 & 2.05 & 4.63 & 3.34\\

        \hline
         \multirow{3}{*}{COM} & 2 & 1.83 & \textbf{4.33} & 2.00 & 4.45 & 3.23 \\

           & 4 & 1.84 & \textbf{4.33} & 2.02 & 4.39 & 3.21\\

           & 32  & \textbf{1.78} & 4.36 & \textbf{1.96} & \textbf{4.33} & \textbf{3.15}\\ \hline
    \end{tabular}

\end{table}
Although we focus our analysis on the 
online scenario, we first tested our models in the offline setting as a sanity check of our pipeline. Results with DIR and COM with different S4 dimensions are reported in Table \ref{table: offline}.
We observe that an S4former using the COM approach achieves the state-of-the-art at offline ASR on Librispeech (1.96\%/4.33\% on test clean/other, 119M params, Table \ref{table: offline}), on par with a larger model from a recently proposed S4-augmented encoder model (1.91\%/4.36\%, 140M params) ~\cite{fathullah2023multihead}. Our Conformer implementation is also on par with published results \cite{gulati2020conformer} (ours: 2.12\%/4.63\%; reported: 2.1\%/4.3\%).

\begin{table}[t]
    \centering
        \caption{WER (\%) of online ASR 
        using the drop-in replacement (DIR) approach.}
    \label{table: DIR results}
    \begin{tabular}{|c|c|c|c|c|c||c|}
    \hline
         \multirow{2}{*}{Scheme} & \multirow{2}{*}{$N$} & dev & dev & test & test & test  \\
          & & clean & other & clean & other & avg.\\
         \hline
         \multirow{3}{*}{S4D-Real} & 2 & \textbf{3.76} & \textbf{9.21} & \textbf{4.18} & \textbf{8.77} & \textbf{6.48}\\

           & 4 & 3.97 & 9.44 & 4.38 & 9.08 & 6.73\\

           & 8 & 3.99 & 9.58 & 4.35 & 9.14 & 6.75\\
        \hline
         \multirow{3}{*}{S4D-Lin} & 2 & 3.90 & 9.53 & 4.40 & 9.23 & 6.82 \\

           & 4 & 3.96 & 9.47 & 4.25 & 9.06 & 6.66\\

           & 8  & 3.96 & 9.37 & 4.29 & 8.97 & 6.63 \\ \hline
    \end{tabular}

\end{table}

\subsection{Online ASR}
\vspace*{-1ex}

We now focus on the online setting, which is a difficult yet important use case not examined by previous research on encoders with S4s. We provide results from exhaustive ablation studies on S4 settings for each of the three approaches. All models, Conformers and S4formers included, have $119 \pm 1M$ trainable parameters. Since our baseline Conformer has hyperparameters tuned for offline but not online ASR, we performed a sweep of its convolution kernel size (from 2 to 64) and used the best setting (4) for comparison with S4formers. All results are summarized in Table \ref{table: summary}.


\subsubsection{Drop-in replacement (DIR)}\vspace*{-1ex}

Results from ablation studies using the DIR approach are shown in Table \ref{table: DIR results}. On test sets, the most performant S4former achieves $4.18\%/8.77\%$, on par with but no better than our tuned Conformer ($4.15\%/8.70\%$). With this approach, most effective S4 to use has the smallest $N$ that we tested. This is somewhat surprising, as previous work on S4s for ASR \cite{gupta2022diagonal} used much larger S4 dimension (64), which was in part motivated by theoretical results showing that S4s with diagonal weights are equivalent to those with non-diagonal weights at infinite S4 dimension (further discussion in \cite{gupta2022diagonal}). Our results indicate that for online ASR, a small-$N$ S4 is the most effective, despite having fewer parameters.

\subsubsection{Combination (COM)}\vspace*{-1ex}

We next test stacking the S4 and convolution to see whether they can act complimentarily. We first sweep through different convolution sizes (Table \ref{table: pre ssm conv}). Adding the convolution resulted in a substantial WER reduction (relative 4-7\%) compared to having the S4 alone (DIR), improving test set WER from 4.38\%/9.08\% to 4.16\%/8.67\%. However, our experiments suggest that for the combination to be effective, kernel size of the convolution needs to be substantially reduced. Using a large-ish kernel size (16), the resultant model in fact performs worse (S4former built with COM gives 4.85\%/9.91\%; Conformer using this size gives 4.46\%/9.46\%). 

We next tested different S4 settings in the COM approach. In these experiments, we fixed the convolution kernel size to 2 and varied S4  parameterization and S4 dimension (Table \ref{table: recurrent weights}). We found that, as in the DIR approach, the most effective S4 here is S4D-Real with a small dimension. 

\begin{table}[t]
\small
    \caption{WERs(\%) of online ASR from encoders using different pre-S4 convolution kernel sizes.  Both DIR and COM used the S4D-Real parameterization scheme with $N=4$ (using other $N$ worsened performance). Note that in COM models, the conv. size only describes the convolution kernel size; the S4s they use have unlimited left context.}
    \label{table: pre ssm conv}
    \centering
    \begin{tabular}{|c|c|c|c|c|c|c||c|}
    \hline
         \multirow{2}{*}{Appr.} & conv. & \multirow{2}{*}{$N$} & dev & dev & test & test  & test\\
           & size & & clean & other & clean & other & avg. \\
           \hline
         \multirow{2}{*}{DIR} &  NA & 2 & 3.76 & 9.21 & 4.18 & 8.77 & 6.48\\
          & NA & 4 & 3.97 & 9.44 & 4.38 & 9.08 & 6.73 \\
        \hline
         \multirow{4}{*}{COM} &  2 & 4 & \textbf{3.67} & \textbf{9.10} & \textbf{4.16} & \textbf{8.67} & \textbf{6.42}\\

          &  4 & 4 & 3.75 & 9.46 & 4.27 & 8.93 & 6.60\\

          &  8 & 4 & 3.75 & 9.35 & 4.17 & 8.98 & 6.58\\

          &  16 & 4 & 4.40 & 10.19 & 4.85 & 9.91 & 7.38\\
    \hline
    \end{tabular}
\end{table}

\begin{table}[]
    \centering
    \caption{WERs(\%) of online ASR from S4formers augmented with the COM approach using different recurrent weights ($\bm{A}$). All models use a pre-S4 convolution with kernel size 2.}
    \label{table: recurrent weights}    
    \begin{tabular}{|c|c|c|c|c|c||c|}
    \hline
         \multirow{2}{*}{Scheme} & \multirow{2}{*}{$N$} & dev & dev & test & test & test \\
          &  & clean & other & clean & other & avg. \\
         \hline
         S4D-Real &  2 & 3.69 & 9.14 & \textbf{4.01} & \textbf{8.53} & \textbf{6.27}\\
         S4D-Real &  4 & \textbf{3.67} & 9.10 & 4.16 & 8.67 & 6.42\\
         S4D-Real &  8 & 3.85 & 9.29 & 4.29 & 8.97 & 6.63 \\
         S4D-Real &  16 & 3.70 & 9.19 & 4.27 & 8.74 & 6.51\\
         S4D-Real &  32 & 3.77 & 9.00 & 4.24 & 8.63 & 6.44\\
         S4D-Real &  64 & 3.70 & \textbf{8.99} & 4.18 & 8.77 & 6.48\\
                 \hline
         S4D-Lin &  4 & 3.71 & 9.21 & 4.25 & 8.86 & 6.56\\

         S4D-Lin &  32 & 3.71 & 9.27 & 4.19 & 8.93 & 6.56\\
                 \hline
    \end{tabular}
    \vspace*{-1ex}

\end{table}


\subsubsection{Reparameterization (REP)}\vspace*{-1ex}

We next considered the final approach, where the S4 is used to reparameterize a finite-size convolution kernel. In all experiments, we use S4s with S4D-Real and dimension 4. To parameterize a convolution kernel of size $L$ using an S4, we compute the kernel as
\begin{equation}
    \tilde{\bm{K}}(L) = [\bm{C} \bar{\bm{B}}, \bm{C} \bar{\bm{A}} \bar{\bm{B}}, ..., \bm{C} \bar{\bm{A}}^{L-1} \bar{\bm{B}}].
\end{equation}
We also note that, while in principle one could stack one convolution component using an S4-parameterized kernel $\tilde{\bm{K}}(L)$ and another one using a conventional kernel, we found that such stacking resulted in performance on par with or worse than 
that of COM (results not presented here). %
As shown in Table \ref{table: truncation}, the best REP model achieves $4.09\%/8.66\%$ on test sets, slightly worse than the best COM model but also better than the tuned Conformer ($4.15\%/8.70\%$), despite having the same kernel size and left context. Compared to DIR or COM, the advantage of this approach is that it does not modify the forward pass of the encoder at test time. The model simply needs to compute and cache the finite-size kernel $\tilde{\bm{K}}(L)$ and perform convolution with it as a vanilla Conformer.

\begin{table}[]
    \caption{WERs(\%) of online ASR from encoders augmented using the reparameterization (REP) approach.}
    \label{table: truncation}
    \centering
    \begin{tabular}{|c|c|c|c|c|c||c|}
    \hline
         \multirow{2}{*}{Appr.} & left & dev & dev & test & test & test \\
          &   cont. & clean & other & clean & other & avg.  \\
         \hline
         \multirow{4}{*}{REP} &  $\infty$ & 3.97 & 9.44 & 4.38 & 9.08 & 6.73 \\
         
           &  32 & 3.86 & 9.35 & 4.21 & 8.97 & 6.59 \\
         
           &  16 & 3.79 & 9.33 & 4.26 & 8.95 & 6.61 \\
         
           &  8 & \textbf{3.75} & \textbf{9.02} & \textbf{4.09} & \textbf{8.66} & \textbf{6.38} \\

           &  4 & 3.80 & 9.21 & 4.18 & 8.83 & 6.51 \\

           &  2 & 3.79 & 9.04 & 4.17 & 8.71 & 6.44 \\ \hline
         \end{tabular}
\end{table}





\subsection{Summary} \vspace*{-1ex}
In Table \ref{table: summary}, we summarize WERs from the most performant S4former using each of the approaches. The Conformer shown here uses the best convolution kernel size we found (size 4). Compared to this carefully tuned Conformer, COM and REP still manage small improvement on test sets, achieving WERs of $4.01\%/8.53\%$ and $4.09\%/8.66\%$, respectively. Our key empirical takeaways are
\begin{itemize}
    \item Compared to Conformers with tuned convolution, S4formers achieve a small but consistent WER reduction in online ASR.
    \item In all settings, S4D-Real is more effective than S4D-Lin.
    \item In online ASR, S4s with very small $N$ (2-4) have the best performance. Larger $N$ is more effective for offline ASR.
    \item How S4s are incorporated into the encoder matters. We found COM $ > $ REP $ > $ DIR.
\end{itemize}

\begin{table}[t]
\small
    \centering
        \caption{Summary of online ASR on Librispeech. All results are from our implementations. The Conformer uses the best setting for online ASR that we found (kernel size 4).}
    \label{table: summary}
    \begin{tabular}{|c|c|c|c|c|c||c|}
    \hline
         \multirow{2}{*}{Model} & \multirow{2}{*}{Appr.} & dev & dev & test & test & test\\
          &   & clean & other & clean & other & avg. \\
         \hline
         Conformer  & NA & \textbf{3.74} & \textbf{9.01} & 4.15 & 8.70 & 6.43\\
         \hline
          & DIR & 3.76 & 9.21 & 4.18 & 8.77 & 6.48\\
         S4former & COM &  3.69 & 9.14 & \textbf{4.01} & \textbf{8.53} & \textbf{6.27} \\
          & REP & 3.75 & 9.02 & 4.09 & 8.66 & 6.38 \\
         \hline
    \end{tabular}
\end{table}

\section{Discussion}\vspace*{-1ex}

We systematically investigated ways to augment encoder models for online ASR with S4s. We proposed two new approaches, COM and REP, both of which resulted in models performing at the state-of-the-art. The success of REP is somewhat surprising, as traditionally S4s are thought to excel at modeling long-range dependencies \cite{gu2021combining, gu2021efficiently, gu2022parameterization}, yet the REP approach prevents them from doing so. This suggests that even for short-range dependencies, S4s can have advantages over simple convolution, pointing to the need of further theoretical research.


\vfill\pagebreak
\bibliographystyle{IEEEbib}
\bibliography{refs.bib}

\end{document}